\pdfoutput=1
\documentclass[11pt]{article}

\usepackage[]{EACL2023}

\usepackage{times}
\usepackage{latexsym}

\usepackage[T1]{fontenc}
\usepackage[utf8]{inputenc}

\usepackage{microtype}
\usepackage{tabularx}
\usepackage{multirow}
\usepackage{graphicx}
\usepackage{booktabs}
\usepackage{amsmath}
\usepackage{arydshln}

\newcommand{\modelname}{\textsc{EveLink}}

\newcommand{\eventlink}{Event Linking}
\newcommand{\entitylink}{entity linking}
\title{
Event Linking: Grounding Event Mentions to Wikipedia}

\author{Xiaodong Yu$^{1}$ \enspace\enspace\enspace Wenpeng Yin$^{2}$ \enspace\enspace\enspace Nitish Gupta$^{3}$ \enspace\enspace\enspace Dan Roth$^{1}$ \\
$^{1}$University of Pennsylvania \enspace\enspace $^{2}$Penn State University \enspace\enspace $^{3}$Google Research\\
\tt \{xdyu, danroth\}@seas.upenn.edu\\
\tt wenpeng@psu.edu\\
\tt gnnitish@gmail.com}

\begin{document}
\maketitle
\begin{abstract}
Comprehending an article requires understanding its constituent events. However, the context where an event is mentioned often lacks the details of this event. A question arises: how can the reader obtain more knowledge about this particular event in addition to what is provided by the local context in the article?

This work defines \eventlink, a new natural language understanding task at the event level. Event linking tries to link an event mention appearing in an article to the most appropriate Wikipedia page. This page is expected to provide rich knowledge about what the event mention refers to. To standardize the research in this new direction, we contribute in four-fold. First, this is the first work in the community that formally defines the \eventlink~task.
Second, we collect a dataset for this new task. Specifically, we automatically gather the training set from Wikipedia, and then create two evaluation sets: one from the Wikipedia domain, reporting the in-domain performance, and a second from the real-world news domain, to evaluate out-of-domain performance. %In specific, 
%we first gather training set automatically from Wikipedia, then create two evaluation sets: one from the Wikipedia domain, reporting the in-domain performance; the other from the real-world news domain, testing the out-of-domain performance. 
%
Third, we retrain and evaluate two state-of-the-art (SOTA) entity linking models, showing the challenges of event linking, and we propose an event-specific linking system,  \modelname, to set a competitive result for the new task. 
Fourth, we conduct a detailed and insightful analysis to help understand the task and the limitations of the current model. 
Overall, as our analysis shows, \eventlink~is a challenging and essential task requiring more effort from the community. \footnote{Data and code are available here: \url{http://cogcomp.org/page/publication_view/996}.}
\end{abstract}

\section{Introduction}

\begin{figure}[t]
    \centering
    \includegraphics[width=0.4\textwidth]{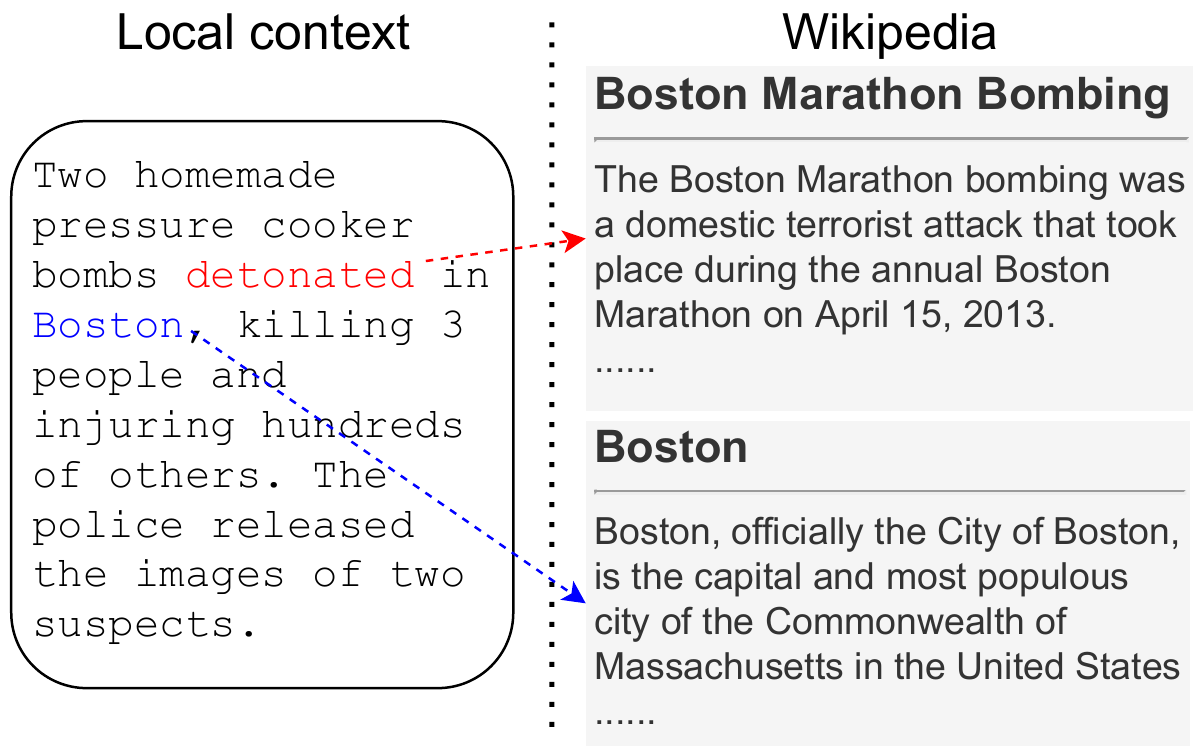}
    \caption{Examples of Event linking and Entity linking. The left side is the local context, and the right side contains Wikipedia pages. Entity linking model connects the entity ``\textcolor{blue}{Boston}'' to the Wikipedia page ``Boston'', while event linking model links the event ``\textcolor{red}{detonated}''  to the Wikipedia page ``Boston Marathon Bombing'', which is more relevant to the local context. }
    \label{fig:boston}
\end{figure}

Grounding is a process of disambiguation and knowledge acquisition, and is an important task for natural language understanding. Entity linking, grounding entity mentions to a knowledge base (usually Wikipedia)
\citep{BunescuPa06, MihalceaCs07, ratinov-etal-2011-local, gupta-etal-2017-entity, wu-etal-2020-scalable},  has been shown important in natural language understanding tasks, such as question answering, recommendation system, dialogue generation. Despite the significant progress brought by \entitylink, we argue that grounding entities may not provide enough background knowledge that is often needed to support %for correct
text understanding. Consider the example, Figure \ref{fig:boston}; an \entitylink~model will link the entity ``Boston'' to the Wikipedia page ``Boston'' which introduces the history and culture of the city Boston. The information we can get from the page ``Boston'' is irrelevant to the local context. To really help understand this sentence, we need to link the event centered by the verb ``detonated'' to the Wikipedia page ``Boston Marathon Bombing''. We call this process that grounds events \textbf{\eventlink}.

In this paper, we formulate this \eventlink~task for the first time, analyze the difference and challenges of the new task, and carefully design a benchmark dataset for this task. We automatically collect training data from the hyperlinks in Wikipedia, and create two evaluation sets to evaluate both in-domain and out-of-domain performance. For in-domain evaluation, the test data is also from hyperlinks in Wikipedia. To avoid models from overfitting, the test data is balanced with hard cases and easy cases determined by whether the event is seen in the training and by the similarity between the surface forms of event mentions and Wikipedia titles. For out-of-domain evaluation, we annotate real-world news articles across 20 years collected from New York Times. Considering the sparsity of events existed in Wikipedia, we also add ``Nil'' annotation to the test data, indicating that  those events  do not exist in Wikipedia, therefore, the model needs to tag them as ``Nil''. 

Technically, we come up with an event linking model \modelname~that uses the entities in the local context as arguments of the event structure to better present the event mention. \modelname~outperforms two SOTA entity linking models BLINK \cite{wu-etal-2020-scalable} and GENRE \cite{DeCao2021AutoregressiveER}, and achieves strong performance on the event linking test set, especially on seen events and easy cases, and a detailed error analysis shows the difficulties of the new task and the limitation of the current model.
% \modelname~consists of two steps. The first step ``candidate generation'' uses a bi-encoder to narrow down the candidate space efficiently; the second step ``candidate ranking'' uses a more advanced cross-encoder to derive the matching degree between an event mention and a Wikipedia title. Both steps rely on a novel representation learning approach, which is our main technical contribution, for event mentions as well as Wikipedia titles.  
% \modelname~achieves strong performance on the event linking test set, especially on seen events and easy cases, and a detailed error analysis shows the difficulties of the new task and the limitation of the current model.

 To conclude, our contributions are four-fold: (i) We formulate the task \eventlink. (ii) We collect training data for this task, and design both in-domain and out-of-domain test data, with a balanced ratio of hard cases and easy cases to ensure the dataset quality. (iii) Our proposed approach \modelname~shows promising performance in experiments, which sets a competitive result for future works. (iv) Our in-depth analysis provides a better understanding of this new problem, the challenges in different domains, and the new approach.

\section{Grounding Events in Wikipedia %Problem Definition
}
\label{sec:def}

Given an article and an event mention $m$ in it, event linking tries to find a title $t$, from \textbf{all} the English Wikipedia titles (around 5m titles), to provide the best explanation of $m$. Event mention is defined as verb or nominal that refers to an event. A correct title is defined as follows: as long as a Wikipedia page is about this event, or any subsection of the page introduces this event, we regard its title as the correct one. In this paper, all the models assume gold event mentions are given. For each event mention, a system is expected to label it with the correct Wikipedia page or a ``Nil'' tag if the event does not exist in Wikipedia. Accuracy is adopted as the official evaluation metric.
\paragraph{\eventlink~vs. Entity Linking.} Relatedness: (i) They both link an object (event/entity) from an article to Wikipedia; (ii) Some events, such as ``World War II'', are entities; in this case, two tasks are the same. 
Distinctions:  (i) Entities are mostly consecutive text spans. Events, in contrast, are more structured objects, consisting of a trigger and a couple of arguments. An event trigger is mostly a general verb, which may not refer to a specific event by its own without knowing event arguments. More complex structures in events make event linking a more challenging task and require a deeper understanding of the local context; (ii) Unlike entities with a large coverage in Wikipedia, many events do not have a record in Wikipedia. Considering the sparsity, we require models to tag event mentions that do not exist in Wikipedia as ``Nil''.   
\paragraph{Why \eventlink?} Except for some events that are also entities, generally speaking, events are information units of larger granularity. As shown in Figure \ref{fig:boston}, a better comprehension of events, such as through linking to Wikipedia, is expected to facilitate the text understanding more. 

\paragraph{Challenges specific to \eventlink.} 

\begin{figure}[t]
    \centering
    \includegraphics[width=0.4\textwidth]{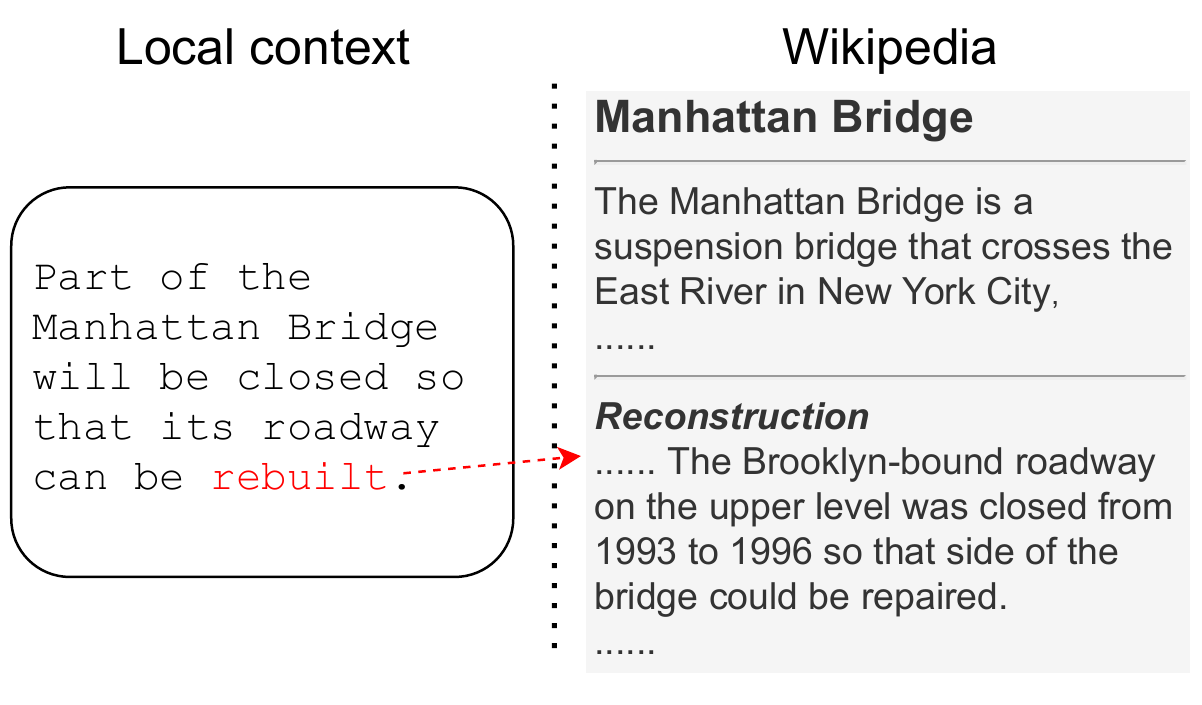}
    \caption{Example of event mentions that only exist in the subsection of a Wikipedia page. The event ``rebuilt'' does not have its own page, but is mentioned in the subsection of the page ``Manhattan Bridge''.}
    \label{fig:bridge}
\end{figure}

\begin{figure}[t!]
    \centering
    \includegraphics[width=0.4\textwidth]{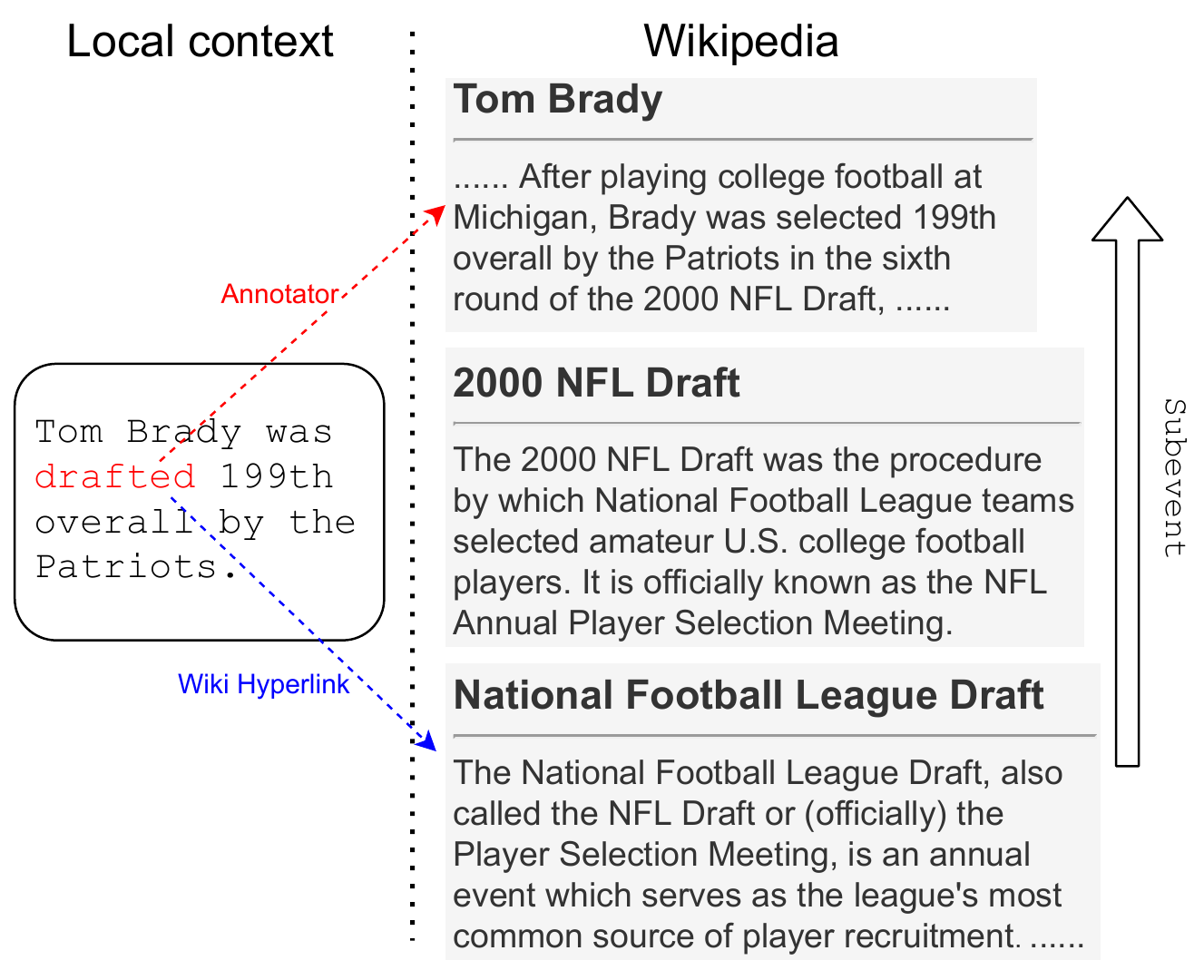}
    \caption{Example of hierarchical events. Event ``draft'' of Tom Brady is mentioned in the page ``Tom Brady'', and is also a sub-event of ``2000 NFL Draft'', which is again a sub-event of ``National Football League Draft''. }
    \label{fig:draft}
\end{figure}

(i) The correct title for some event mentions may not be unique. The same event could be introduced in several pages. For example, ``Invasion of Poland'' and ``Occupation of Poland (1939–1945)'' both introduce the event that German Army invaded Poland in 1939. How to decide the ground truth set and how to evaluate in this situation are not trivial.

(ii) Events may only exist in the subsection of the Wikipedia page. Only a limited number of famous events have their own pages, while many other relatively infamous events only exist in the subsection of some pages. Considering the example in Figure \ref{fig:bridge}, the event ``rebuilt'' of the Manhattan Bridge does not have its own Wikipedia page, but it is mentioned in the subsection ``Reconstruction'' of the page ``Manhattan Bridge''. Linking these events requires a model to understand the whole page instead of just encoding the first paragraph.

(iii) Events have a hierarchical structure. Events at larger granularity consist of many sub-events, and these sub-events may have their own Wikipedia pages, or just be mentioned in the pages of the large events. Ideally, the model should always link the event mention to the most appropriate page. 
If the sub-event page exists, then link to the sub-event page. Otherwise, link to the page of the large event. 
However, the term  ``appropriate'' here could be unclear because of the event hierarchy. As Figure \ref{fig:draft} shows, the Wikipedia page ``Tom Brady'' is most specific to  the event ``drafted''. On the other hand, draft of ``Tom Brady'' is a sub-event of  ``2000 NFL Draft'', which is further a sub-event of  ``National Football League Draft''. Annotators prefer to link this event to ``Tom Brady'', while Wikipedia hyperlinks link the event to ``National Football League Draft''. The hierarchy of events makes the standard of the correct title inconsistent.

\section{Related Work}

\textbf{Entity Linking.} As described in the previous section, entity linking has been extensively studied for many years \citep{BunescuPa06, MihalceaCs07, ratinov-etal-2011-local, gupta-etal-2017-entity, wu-etal-2020-scalable, DeCao2021AutoregressiveER}. Though both of entity linking and event linking could be regarded as a task linking document contents to a knowledge base, we argue that entity linking is more about linking text span, while event linking is more about linking an event structure, centered by a predicate, which is more challenging because the predicate span is usually a general verb. In the experiment section, we show that just retraining the entity linking model on event linking data without considering the event structure does not perform well. \citet{humeau2019poly} and \citet{wu-etal-2020-scalable} use a bi-encoder/cross-encoder architecture to train the candidate generation/ranking model respectively for entity linking. Considering the structure of events that entities do not have, \modelname~extends their model by adding structure information to the event mention representation.\\
\textbf{``Event Linking''.}~ We note that the term ``event linking'' has been once used in the literature \cite{DBLPmanHHC12,krvent}. However, these works are essentially performing cross-document event coreference: determine if a given event mention refers to another event mention (in the same or another document). 
We, on the other hand, link an event mention to a Wikipedia concept with a different purpose: acquiring external knowledge about the event which is often beyond what we can obtain from the local context. Our definition of event linking can not only improve the understanding of the article, but also pave the way for the intensively-studied event coreference and other event relation identification problems.\\
% \textbf{Event Representation.} \citet{vyas-ballesteros-2021-linking} use similar methods to add entity attributes to the entity representation as our method of adding entities to the event representation. The difference is that we also add entity type information to the representation by using special tokens to indicate the start and end of entities in different types.\\
\textbf{Data.} \citet{eirew-etal-2021-wec} collect training data from Wikipedia hyperlinks for event coreference, while we use similar methods to collect data for event linking. In this work, we use the FIGER type of the title to find event titles, while \citet{eirew-etal-2021-wec} use the Wikipedia infobox. There also exists some other event knowledge bases, such as EventKG \cite{gottschalk2018eventkg}. Because we use hyperlinks in Wikipedia as training data resource, and we do not limit the candidate space to be event titles only, in this work we only focus on linking event mentions to Wikipedia, and the candidate space is all the Wikipedia titles.

\section{Data Construction}

We collect training data and in-domain test data from Wikipedia automatically, and manually annotate a test set in the news domain for out-of-domain evaluation purpose. Table \ref{tab:examples} lists some data examples, and Table \ref{tab:wikistat} shows detailed statistics.

    \subsection{Wikipedia} 
    \label{sec:wikidata}

    \begin{table*}[t]
    \small
    \centering
    \begin{tabular}{l|l|c}
    \toprule
     & Event mention in local context & Wikipedia title \\
    \midrule
    \multirow{6}{*}{Wiki} & At the start of the \textcolor{red}{wartime 1940s} , he had four releases. & World War II \\
    \cdashline{2-3}
    & Henry Louis Gates, a black Harvard University professor \textcolor{red}{who} & \multirow{3}{*}{Henry Louis Gates arrest controversy}\\
    & \textcolor{red}{was arrested} after police mistakenly thought he was breaking into \\
    & his own home in Cambridge, Massachusetts. \\
    \cdashline{2-3}
    & Ibrox hosted four Scotland games in the first phase, starting with a & \multirow{2}{*}{1994 FIFA World Cup qualification}\\
    & \textcolor{red}{1994 World Cup qualifier} against Portugal in October 1992.\\
    \midrule
    % \multirow{6}{*}{NYT} & The move is in retaliation for \textcolor{red}{efforts} by China to get around & \multirow{2}{*}{China–United States trade war} \\
    % & American limits on imports by shipping goods through other nations\\
    \multirow{5}{*}{NYT} & The Nets \textcolor{red}{offered} Sam Cassell and a first-round draft pick for Marbury. & Sam Cassell \\
    % & American limits on imports by shipping goods through other nations\\
    
    \cdashline{2-3}
    & A man who \textcolor{red}{killed} his former wife, a bartender and a cook in 1984 & \multirow{2}{*}{Godinez v. Moran}\\
    & was executed by injection early today. \\
    \cdashline{2-3}
    % & The Department of Defense has identified 2,316 American service & \multirow{2}{*}{Iraq War}\\
    % & members who have \textcolor{red}{died} since the start of the Iraq war. \\
    % \cdashline{2-3}
    % & The Nets \textcolor{red}{offered} Sam Cassell and a first-round draft pick for Marbury. & Sam Cassell\\
    % \cdashline{2-3}
    & A 45-year-old fashion photographer was shot and \textcolor{red}{killed} in his West & \multirow{2}{*}{Nil}\\
    & Village apartment yesterday morning, the police said.\\
    \bottomrule
    \end{tabular}
    \caption{Data examples. The upper part is data collected from Wikipedia hyperlinks. The lower part is annotated New York Times (NYT) paragraphs. Event mentions are highlighted in red.}
    \label{tab:examples}
\end{table*}

    \begin{table}[t]
        \small
        \centering
        \begin{tabular}{lcc|cc}
        \toprule
        & Train & Dev & \multicolumn{2}{c}{Test}\\
         & Wiki & Wiki & Wiki  & NYT \\
        \midrule
        Verb & 33,213 & 8,346 & 9,633 & 1,319\\
        \enspace Seen Event & - & 1,814 & 2,913 & 0\\
        \enspace Unseen Form & - & 2,585 & 3,828 & 75\\
        \enspace Unseen Event& - & 3,947 & 2,892 & 435\\
        \enspace Nil & - & - & - & 809\\
        Nominal & 33,213 & 8,346 & 9,633 & 443\\
        \enspace Hard & - & 4173 & 4817 & 244\\
        \enspace Easy & - & 4173 & 4817 & 15\\
        \enspace Nil & - & - & - & 184\\
        Total & 66,426 & 16,692 & 19,266 & 1,762\\
        \bottomrule
        \end{tabular}
        \caption{Wikipedia and New York Times (NYT) data statistics. NYT is only for testing.} 
        \label{tab:wikistat}
    \end{table}

    We first collect all hyperlinks $ (hypertext,\ title) $ in Wikipedia text, which links a hypertext to a Wikipedia title. Then, we map the FreeBase type of  Wikipedia titles to FIGER types \citep{LingWe12}, and all titles with a type ``Event'' are regarded as event titles. All the hypertexts linked to these event titles are regarded as event mentions.
    
    Because same event mentions in one Wikipedia page are hyperlinked only once, and editors tend to hyperlink more nominal mentions than verb mentions, verb mentions are highly limited in Wikipedia. To balance the size of verbs and nominals, we use SpaCy Part-of-Speech model\footnote{\url{https://spacy.io/usage/linguistic-features##pos-tagging}} to keep all verb mentions, and sample the same size of nominals. To prevent models from overfitting, we design hard and easy cases for verbs and nominals:
    
    \textbf{Verbs}: We classify each verb mention mainly by whether the surface form (\textsc{S}) of the verb is seen in training data, and whether the gold event title (\textsc{T}) is seen in training data. If both \textsc{S} and \textsc{T} are seen in training data, we call it \textbf{Seen Event}. If \textsc{T} is seen in training data, but  \textsc{S} is new, we call it \textbf{Unseen Form}. If \textsc{T} is never seen in training data, we call it \textbf{Unseen Event}. Under this setting, ``Seen Event" is regarded as easy cases, and the other two are hard cases.
    % We require all the models to report performance on all the categories. 
    Because of the limited size of verb mentions, all the event titles with fewer than or equal to 5 verb mentions are used as ``Unseen Event".
    
    \textbf{Nominals}: We classify each nominal mention mainly by its surface form similarity to the gold title. 
    % For example, the gold title of the event mention ``World War II" is also ``World War II", which is a easy case. 
    We calculate the Jaccard similarity between the nominal mention and the gold title by taking 3 grams of the surface form. If the similarity is lower than 0.1, we think it is a \textbf{hard nominal}; otherwise, it is an \textbf{easy nominal}. Then we sample same numbers of hard and easy cases. 
    
    % Data samples are shown in the upper part of Table \ref{tab:examples}. Detailed statistics are shown in Table \ref{tab:wikistat}. 

    \subsection{New York Times}
    
    % \begin{table}[t]
    %     \small
    %     \centering
    %     \begin{tabular}{lc}
    %     \toprule
    %      & Test \\
    %     \midrule
    %     Verb & 1,319\\
    %     \enspace\enspace Seen Event & 0 \\
    %     \enspace\enspace Unseen Form & 75 \\
    %     \enspace\enspace Unseen Event & 435 \\
    %     \enspace\enspace Nil & 809 \\
    %     Nominal & 443 \\
    %     \enspace\enspace Hard & 244 \\
    %     \enspace\enspace Easy & 15 \\
    %     \enspace\enspace Nil & 184 \\
    %     Total & 1,762 \\
    %     \bottomrule
    %     \end{tabular}
    %     \caption{New York Times data statistics. NYT is only for evaluation purpose.} 
    %     \label{tab:nytstat}
    % \end{table}

    We sample 2,500 lead paragraphs from The New York Times Annotated Corpus \citep{nyt2008}, which contains New York Times articles from 1987 to 2006. We first use an off-the-shelf verb and nominal SRL model\footnote{\url{https://cogcomp.seas.upenn.edu/page/demo_view/SRLEnglish}}
    to extract event mention candidates, and then we use Amazon Mechanical Turk to annotate the corresponding Wikipedia title of the predicted mention candidates. To ensure the quality of the annotation, we design our annotation process in two rounds: \\
    \textbf{First round.} Annotators need to answer whether they think the predicted mention is an event or not. If they think it is an event, then they need to find the corresponding Wikipedia title, otherwise submit ``Nil''. Each mention is annotated by three annotators. If all of them submit ``Nil'', we include this event mention as a ``Nil'' example in the final test data. \textit{To prevent annotators from simply submitting ``Nil'', 10\% of the event mentions are the relatively easy cases from the Wiki data and we know their answers.  We randomly insert them into the input data for AMturk (i.e., annotators are unaware of that) to evaluate the accuracy of the annotator. Only the annotation from  annotators with an accuracy higher than 90\% will be accepted.} \\
    \textbf{Second round.} This round verifies the annotated results in the first round. 
    % The intuition here is that it is hard to find the correct Wikipedia title, especially for those events annotators are not familiar with because it is unclear whether you have not found the title, or it is just a ``Nil''. Also, many events are only introduced in the subsection of a Wikipedia page, which makes the first round annotation harder. However, verification is a much simpler task. 
    Each mention with the annotated title is verified by another three annotators. They need to read the page, and figure out whether it introduces the mention.  If the majority vote for ``yes'', we include it in the final test data. Because of majority voting, some annotations that not all the annotators agree would be included. The inter-agreement is 63.74 Fleiss' kappa.

    \subsection{Domain Analysis} \label{sec:domain}
    
    Event linking in the news domain is more challenging than that in the Wikipedia domain because of the following reasons:
    
    (i) News articles describe an event at a different granularity as how Wikipedia does, usually with more details. For example, here is a piece of news about ``Iraq\_War'': "A contractor working for the American firm Kellogg Brown \& Root was \textcolor{red}{wounded} in a mortar attack in Baghdad." The event ``wounded'' here is a very small event in Iraq War, but it is what daily news would report. On the other hand, the event mention that links to ``Iraq\_War'' in Wikipedia domain is: "When touring in Europe, the US \textcolor{red}{went to war} in Iraq." The different granularity in representing events makes the task slightly different in two domains. Event linking in Wikipedia domain is more like event coreference, while event linking in news domain is mixed with more sub-event relation extraction.
    
    (ii) As analyzed in Section \ref{sec:def}, event linking is challenging because some event mentions may only exist in the subsection of the correct page, and the correct title is not consistent because of the event hierarchy. However, these problems mainly happen in the news domain. First of all, the Wikipedia hyperlinked mentions usually have their own pages instead of just existing in subsections. In news domain, we annotate events that only exist in subsections of a Wikipedia page. Second, in Wikipedia domain, the gold title of same event mentions is usually consistent. For example, all of the event mentions ``drafted'' of football players link to ``National Football League Draft'' instead of the page of the specific player. However, the annotation standard of NYT is not always consistent with Wikipedia hyperlinks. For example, annotators would link event mentions about sports player draft to the page of the specific player instead of the general concept page ``National Football League Draft''. These problems make data annotation and model evaluation in news domain very challenging.

    Because of the reasons claimed above, we think that, for some cases in news domain, the correct answer is multiple titles instead of just one title. Ranking the annotated title to the second place may be because the top one is also correct. To relax the evaluation metric here, for news domain, we also report the number of Accuracy@5, which means that if the annotated title is ranked in the top 5 candidates, we think it is correct.

    \begin{table*}[ht]
        \small
        \setlength{\tabcolsep}{5pt}
        \centering
        \begin{tabular}{lcccccccccc}
            \toprule
            Models & \multicolumn{4}{c}{Verb} & & \multicolumn{3}{c}{Nominal} & & Verb + Nominal \\ 
            \cline{2-5}
            \cline{7-9}
             & Seen & Unseen Form & Unseen & Overall & & Hard & Easy & Overall && \\
            \midrule
            % Prior & 87.75 & 30.85 & 21.45 & 57.88 && 38.59 & 90.05 & 60.02 && 59.28\\
            Glove & 23.70 & 16.57 & 14.89 & 18.22 && 3.08 & 84.60 & 43.88 && 30.98 \\
            BM25 & 32.17 & 47.41 & 61.86 & 47.14 && 22.54 & 33.22 & 27.88 && 37.51\\
            BLINK-Entity & 88.91 & 69.64 & 62.31 & 73.37 && 67.93 & 95.20 & 81.57 && 77.42\\
            BLINK-Event & 80.88 & 85.84 & 84.54 & 83.95 && 79.39 & 89.10 & 84.24 && 84.10\\
            GENRE-Entity & 92.41 & 73.48 & 59.13 & 74.90 && 79.84 & \textbf{96.53} & 88.19 && 81.54\\
            GENRE-Event & \textbf{98.80} & 87.30 & 58.85 & 82.24 && 85.34 & 94.64 & 89.99 && 86.12\\
            \midrule
            \modelname & 93.99 & \textbf{92.74} & \textbf{93.91} & \textbf{93.47} && \textbf{89.79} & 95.52 & \textbf{92.65} && \textbf{93.06}\\ 
            \bottomrule
        \end{tabular}
        \caption{Recall on Wikipedia Test. ``Seen'' means both the surface forms of the mention and the gold title are seen in training. ``Unseen Form'' means the surface form of the mention is new, but the gold title is seen in training. ``Unseen'' means that the gold title is unseen in training. BLINK-Entity is the original BLINK model trained on entity linking dataset. BLINK-Event is trained on the new event linking dataset. More details in Section \ref{sec:exp}} 
        \label{tab:wikirecall}
    \end{table*}
    
    \begin{table*}[ht]
        \small
        \setlength{\tabcolsep}{5pt}
        \centering
        \begin{tabular}{lccccccccccc}
            \toprule
            Models & \multicolumn{4}{c}{Verb} & & \multicolumn{3}{c}{Nominal} & & Verb + Nominal \\ 
            \cline{2-5}
            \cline{7-9}
             & Seen & Unseen Form & Unseen & Overall & & Hard & Easy & Overall && \\
            \midrule
            Prior & 62.21 & 2.38 & 1.24 & 38.81 && 34.65 & 85.99 & 61.65 && 54.79\\
            % Cosine Similarity \\
            BLINK-Entity & 64.13 & 48.56 & 45.92 & 52.48 && 46.79 & 88.27 & 67.53 && 60.00\\
            BLINK-Event & 77.72 & 69.78 & 62.72 & 70.06 && 62.59 & 82.29 & 72.44 && 71.25\\
            GENRE-Entity & 75.04 & 57.00 & 44.85 & 58.81 && 65.29 & \textbf{90.91} & 78.10 && 68.45\\
            GENRE-Event & \textbf{95.50} & 73.80 & 45.16 & 71.76 && 72.60 & 88.04 & 80.32 && 76.04\\
            \midrule
            \modelname & 91.21 & \textbf{80.30} & \textbf{78.08} & \textbf{82.93} && \textbf{75.90} & 89.70 & \textbf{82.80} && \textbf{82.87} \\ 
            \bottomrule
        \end{tabular}
        \caption{Accuracy on Wikipedia Test. } 
        \label{tab:wikiaccuracy}
    \end{table*}

\section{Model}
In this section, we propose \modelname \ as the first event linking model. We first introduce the representation of event mentions and event titles in Section \ref{sec:rep}, and then introduce the model architecture in Section \ref{sec:generation}.

\subsection{Event Representation} \label{sec:rep}

A key difference between entity and event is that the context of an entity is more diverse than the context of an event. For example, when the entity ``China'' is mentioned in a sentence, it is unclear what entities or what events probably would also be mentioned together. However, if a verb like ``invade'' is used to represent the event ``Battle of France'' in a sentence, it is very likely that entities like ``Germany'', ``Italy'' and ``France'' will also be mentioned. This shows that an event is defined by its arguments, and these arguments, with a large chance, will also be mentioned in the local context because the verb itself cannot refer to any event. Given this observation, we think that the entities in the local context of the event mention should overlap with the entities in the correct Wikipedia page, and these entities can be used to help the model better represent events. To embed these entities information explicitly to the event representation, we use similar method as how \citet{vyas-ballesteros-2021-linking} embed entity attributes information to the entity representation.\\\\
\textbf{Event mentions}: To represent event mentions in local context, we first use an off-the-shelf Named Entity Recognition model 
% \footnote{Link will be released after anonymity period.} 
\footnote{\url{https://cogcomp.seas.upenn.edu/page/demo_view/NEREnglish}} 
trained on 18-type OntoNotes dataset \citep{weischedel2013ontonotes} to extract the entities around the event. We simply define the context window by 500 characters around the event mention. After predicting all the entities $e_i$ with their type $t_i$, we represent the event mentions by:
    \begin{align}
        r_1 &= [\mathrm{CLS}]\ \mathrm{ctxt}_l\ [\mathrm{M}_s]\ \mathrm{m}\ [\mathrm{M}_e]\ \mathrm{ctxt}_r \\
        r_2 &= [\mathrm{t}_{1_s}]\ \mathrm{e}_1\ [\mathrm{t}_{1_e}]\ \cdots [\mathrm{t}_{n_s}]\ \mathrm{e}_n\ [\mathrm{t}_{n_e}] \\
        r_m &= r_1\ [\mathrm{SEP}]\ r_2\ [\mathrm{SEP}]
    \end{align}
where $\mathrm{m}$, $\mathrm{ctxt}_l$, $\mathrm{ctxt}_r$, $\mathrm{e}_i$ are tokens of event mention, the context on the left of the mention, the context on the right of the mention and predicted entities. $[\mathrm{M}_s]$ and $[\mathrm{M}_e]$ are special tokens to tag the start and end of the event mention. $[\mathrm{t}_{i_s}]$ and $[\mathrm{t}_{i_e}]$ are special tokens to tag the start and end of the entity whose type is $t_i$. $r_m$ is the final representation of event mentions.\\\\
\textbf{Title}: To represent Wikipedia titles, since important entities are already hyperlinked in the page contents, we take the first ten hyperlinked spans as entities, and represent the title by:
    \begin{align}
        r_3 &= [\mathrm{CLS}]\ \mathrm{title}\ [\mathrm{TITLE}]\ \mathrm{description} \\
        r_4 &= \mathrm{h}_1\ [\mathrm{SEP}]\ \mathrm{h}_2\ [\mathrm{SEP}]\ \cdots\ [\mathrm{SEP}]\ \mathrm{h}_n \\
        r_t &= r_3\ [\mathrm{SEP}]\ r_4\ [\mathrm{SEP}]
    \end{align}
where $\mathrm{title}$, $\mathrm{h}_i$ and $\mathrm{description}$ are tokens of the title, hyperlinked spans, and the content of the Wikipedia page. We simply take the first $2,000$ characters as the description. $[\mathrm{TITLE}]$ is the special token to separate the title and the description. $r_t$ is the final representation of Wikipedia titles.

    \begin{table*}[ht]
        \small
        \setlength{\tabcolsep}{5pt}
        \centering
        \begin{tabular}{lcccccccccc}
            \toprule
            Models & \multicolumn{4}{c}{Verb} & & \multicolumn{3}{c}{Nominal} & & Verb + Nominal \\ 
            \cline{2-5}
            \cline{7-9}
             & Seen & Unseen Form & Unseen & Overall & & Hard & Easy & Overall && \\
            \midrule
            Glove & - & 0.00 & 0.70 & 0.60 && 0.00 & 33.33 & 1.66 && 0.94 \\
            BM25 & - & 28.38 & 41.74 & 39.80 && 45.27 & 31.25 & 44.40 && 41.35 \\
            BLINK-Entity & - & 4.00 & 6.67 & 6.27 && 7.79 & 60.00 & 10.81 && 7.80\\
            BLINK-Event & - & 35.14 & 37.39 & 37.06 && 37.45 & 75.00 & 39.77 && 37.97\\
            GENRE-Entity & - & 18.92 & 9.86 & 11.18 && 9.88 & 62.50 & 13.13 && 11.83\\
            GENRE-Event & - & \textbf{56.77} & 17.66 & 23.33 && 21.81 & 31.25 & 22.39 && 23.02\\
            \midrule
            \modelname & - & 52.70 & \textbf{59.40} & \textbf{58.43} && \textbf{51.03} & \textbf{93.75} & \textbf{53.68} && \textbf{56.83}\\ 
            \bottomrule
        \end{tabular}
        \caption{Recall on New York Times data. Because ``Nil'' mentions do not have the Wikipedia title, the Recall is only evaluated on the mentions that exist in Wikipedia.} 
        \label{tab:nytrecall}
    \end{table*}
    
    \begin{table*}[ht]
        \small
        \setlength{\tabcolsep}{5pt}
        \centering
        \begin{tabular}{lcccccccccc}
            \toprule
            Models & \multicolumn{3}{c}{Verb} & & \multicolumn{3}{c}{Nominal} & & \multicolumn{2}{c}{Verb + Nominal}\\ 
            \cline{2-4}
            \cline{6-8}
            \cline{10-11}
              & Unseen Form & Unseen & Overall & & Hard & Easy & Overall && Accu@5 & Accu@1\\
            \midrule
            Prior & 0.00 & 0.00 & 0.00 && 0.00 & 6.25 & 0.39 && 0.52 & 0.13 \\
            BLINK-Entity  & 1.33 & 2.76 & 2.55 && 4.92 & 33.33 & 6.56 && 11.44 & 3.90 \\
            BLINK-Event  & 17.57 & 5.28 & 7.06 && 11.11 & 37.50 & 12.74 && 17.04 & 8.97\\
            GENRE-Entity & 8.11 & 5.73 & 6.08 && 3.29 & 31.25 & 5.02 && 11.83 & 5.72\\
            GENRE-Event & \textbf{39.19} & 8.03 & 12.55 && 7.82 & 31.25 & 9.27 && 23.02 & 11.44\\
            \midrule
            \modelname  & 28.37 & \textbf{13.07} & \textbf{15.29} && \textbf{14.81} & \textbf{43.75} & \textbf{16.60} && \textbf{29.13} & \textbf{15.73}\\ 
            \bottomrule
        \end{tabular}
        \caption{Accuracy on New York Times data without Nil. Only event mentions that exist in Wikipedia are given. Accu@5 means the correct title is ranked top 5. Accu@1 means the correct title is top 1.} 
        \label{tab:nytaccuracy}
    \end{table*}
    
    \begin{table*}[ht]
        \small
        \setlength{\tabcolsep}{5pt}
        \centering
        \begin{tabular}{lcccccccccccc}
            \toprule
            Models &\multicolumn{4}{c}{Verb} && \multicolumn{4}{c}{Nominal} && \multicolumn{2}{c}{Verb+Nominal}\\ 
            \cline{2-5}
            \cline{7-10}
            \cline{12-13}
             & Unseen Form & Unseen & Nil & Overall && Hard & Easy & Nil & Overall && Accu@5 & Accu@1 \\
            \midrule
            BLINK-Entity & 2.7 & 1.15 & 79.85 & 49.51 && 1.23 & 25.00 & 63.04 & 27.77 && 57.26 & 44.04\\
            BLINK-Event &  12.16 & 1.61 & 90.85 & 56.94 && 4.53 & 37.50 & 88.59 & 40.63 && 58.45 & 52.84\\
            \midrule
            \modelname & \textbf{17.57} & \textbf{4.59} & \textbf{93.08} & \textbf{59.59} && \textbf{7.00} & \textbf{43.75} & \textbf{89.67} & \textbf{42.66} && \textbf{59.70} & \textbf{55.33}\\ 
            \bottomrule
        \end{tabular}
        \caption{Accuracy on New York Times data with Nil. We simply predict all the mentions with a probability lower than 50 to Nil.} 
        \label{tab:nytaccuracynil}
    \end{table*}

\subsection{Model Architecture} \label{sec:generation}
Similar to \citet{wu-etal-2020-scalable}, we first use a bi-encoder architecture to efficiently generate candidates, and use a cross-encoder architecture, which requires more computations, to rank the candidates. 

\textbf{Candidate Generation.} We use a bi-encoder architecture to train the candidate generation model. We use two independent BERT transformers \citep{Devlin2019BERTPO} to encode the representation of event mentions $r_m$ and Wikipedia titles $r_t$, and use the output of the two $\mathrm{[CLS]}$ tokens in $r_m$ and $r_t$ as the event mention vector $v_m$ and the title vector $v_t$. Then, we maximize the dot product between the vectors of event mentions $v_m$ and the correct title $v_t$ in a batch with randomly selected negatives. At inference time, representations of all the titles are cached, and for each event mention, we calculate the dot products between its representation and the representation of all the titles, and titles with higher scores will become candidates. 
% We refer more details to \citet{wu-etal-2020-scalable}. 

\textbf{Candidate Ranking.} For each event mention, we take 30 candidates from the candidate generation model as the training data for the ranking model, and use a cross-encoder architecture to train the candidate ranking model. We concatenate the representation of event mentions $r_m$ and titles $r_t$, use one BERT transformer to encode the concatenated representation, and use the output of the $\mathrm{[CLS]}$ token as the final vector $v$. Then we maximize the dot product between the vector $v$ of the correct title and an additional linear layer $W$.
% We refer more details to \citet{wu-etal-2020-scalable}. 

\section{Experiments} \label{sec:exp}
In this section, we evaluate the in-domain performance on Wiki test set and the out-of-domain performance on NYT test set, and conduct an error analysis. Implementation details in Appendix \ref{sec:implement}.
  
% \paragraph{Implementations} We use 4 Nvidia RTX A6000 48GB GPUs for model training and evaluation. For both candidate generation model and candidate ranking model, we train 10 epochs with learning rate $1\mathrm{e}^{-5}$, and use BERT-large-uncased as the pretrained language model \citep{Devlin2019BERTPO}. The maximum tokens of both event mention representation and Wikipedia title representation are 256.

\textbf{Baselines.} Since there is no existing event linking system, we have to compare with previous entity linking systems. In this paper, we mainly compare our system with two SOTA entity linking models BLINK \citep{wu-etal-2020-scalable} and GENRE \citep{DeCao2021AutoregressiveER}. To make a fair comparison, BLINK and GENRE have the following two setups:

\textbf{BLINK/GENRE-Entity}: Since a large portion of event mentions are nominals, which is also a kind of entity, it would be interesting to see how a SOTA entity linking system performs for event linking. Therefore, we test the BLINK/GENRE model pretrained specific to entity linking directly. Please note that the size of entity linking training data is 9 million, which is much larger than the size of event linking training data 66k.

\textbf{BLINK/GENRE-Event}: It adopts the same algorithm with the original BLINK/GENRE system, but is trained on our event linking training set.

For all the experiments, BLINK-Entity retrieves 10 candidates from candidate generation, and both BLINK-Event and \modelname~retrieves 100 candidates from candidate generation. These numbers are tuned on dev data. GENRE is a generation model, which does not use the same pipeline of candidate generation and ranking. We follow the original setting to use the beam search with 5 beams.

Besides SOTA entity linking systems, we also evaluate the performance of BM25, Glove vector cosine similarity between event mention and titles \cite{Pennington2014GloVeGV} and prior distribution. Because event mentions are limited in Wikipedia, to fairly estimate the prior distribution of the event titles, we only evaluate event mentions that appear at least 10 times in Wikipedia. 

% \paragraph{Configuration} For both Wikipedia domain and news domain, in addition to reporting the overall performance, we particularly list; Given Nil or not

\paragraph{In-domain experiment on Wikipedia.}\label{sec:indomain}
    
    We evaluate \modelname~on the Wikipedia test set as the in-domain performance. We report the recall of candidate generation in Table \ref{tab:wikirecall}, and the accuracy of candidate ranking in Table \ref{tab:wikiaccuracy}. As shown in Table \ref{tab:wikirecall} and Table \ref{tab:wikiaccuracy}, \modelname~outperforms baseline models by a large margin, $6.94$ points in Recall and $6.38$ points in Accuracy. \modelname~also achieves a high performance on seen verbs and easy nominals, around 90 accuracy, but a relatively low performance on other hard cases, which leaves a large space for future works to further improve.

\paragraph{Out-of-domain experiment on News.}\label{sec:outofdomain}
    
    We evaluate \modelname~on the NYT test set as the out-of-domain performance. In Table \ref{tab:nytrecall}, we evaluate the recall of candidate generation. Because ``Nil'' mentions do not have correct titles in Wikipedia, we only evaluate the the recall of event mentions that exist in Wikipedia. Though the recall of \modelname~is much higher than the recall of other baseline models (56.83 vs. 37.97), the recall drops significantly compared with the recall on Wikipedia test set (56.83 vs. 93.06). In Table \ref{tab:nytaccuracy}, we evaluate the accuracy on the event mentions that exist in Wikipedia, which is the same setting as the experiments in the Wikipedia domain, and again the accuracy drops significantly from $82.87$ to $15.73$. Even if we accept 5 predictions instead of just one to solve the multiple correct titles problem, the Accuracy@5 is 29.13, which is still low. Detailed error analysis is in Section \ref{sec:analysis}. In Table \ref{tab:nytaccuracynil}, we evaluate the accuracy of all the event mentions, including Nil. Because we do not have Nil examples in training data and development data, we simply predict all the event mentions with probability lower than 50 to ``Nil'', and leave better solutions to future works. GENRE is not tested for Nil mentions because it is unclear how to get its prediction probability.
    
\paragraph{Analysis.}\label{sec:analysis} We wonder following questions:

    \begin{table}[t]
        \small
        \setlength{\tabcolsep}{5pt}
        \centering
        \begin{tabular}{lccc}
            \toprule
            Models & Wiki Test & NYT (no Nil) & NYT\\
            \midrule
            \modelname & 82.87 & 15.73 & 55.33\\
            \midrule
            - type & 81.39 & 11.70 & 55.96\\
            - entities & 71.25 & 8.97 & 52.84\\
            \bottomrule
        \end{tabular}
        \caption{Ablation Study of \modelname} 
        \label{tab:ablation}
    \end{table}

% In this section, we do an analysis for our approach \modelname. 
\noindent
\textbf{$\mathcal{Q}_1$:} Where the gain comes from, compared with the BLINK system?

We do ablation study in Table \ref{tab:ablation}. Explicitly adding entities to the event representation boosts the performance by 10.14 accuracy on Wiki test data and 2.73 accuracy on NYT data. Adding entity types further improves the performance by 1.48 accuracy on Wiki and 4.03 accuracy on NYT data.

\noindent
\textbf{$\mathcal{Q}_2$:} Error patterns of \modelname

We collect several error patterns that are common in both domains, and patterns that are mostly in news domains. Error patterns of both domains:

(i): Repeating events. In the errors, we find many repeating events, like award ceremonies or sports games, that would happen every several years, and the model usually cannot find the correct year of the event if the year is not explicitly mentioned in the context. For example:

    \begin{minipage}{0.93\linewidth}
    {\fontsize{10}{12}\selectfont In 1995, his debut season, Biddiscombe made two appearances, $\cdots$  The following year he earned a \textcolor{red}{Rising Star nomination} for his performance$\cdots$}
    \end{minipage}
In this example, the gold event is ``1996 AFL Rising Star'', and the prediction is ``1998 AFL Rising Star'', though there is a temporal hint (the following year of 1995 is 1996) to indicate that the correct answer should be the award in 1996. There are many similar errors when linking awards or games, which shows that a deeper temporal understanding is necessary for future works.

(ii): Unrelated context. \modelname~replies on the surrounding entities to link the event mentions, however, the context is not always related and surrounding entities cannot help linking. For example: 

    \begin{minipage}{0.93\linewidth}
    {\fontsize{10}{12}\selectfont Returning to his country at the end of the conflict and another \textcolor{red}{begun}, Barinaga rejected an offer from Athletic Bilbao, moving to Real Madrid instead.}
    \end{minipage}
In this example, the gold event is ``World War II'', but the prediction is ``1939–40 La Liga''. All the entities, like ``Barinaga'', ``Athletic Bilbao'' and ``Real Madrid'', are about football, which is unrelated to the war. To link to the correct page, the model needs to know the second conflict of Barinaga's country, which indicates that only using the local context maybe not enough.\\
Error patterns specific to news domain:\\
(i): Subsection events. Some events do not have their own pages, and are only introduced in the subsections of other pages. For example: 

    \begin{minipage}{0.93\linewidth}
    {\fontsize{10}{12}\selectfont The Philippine government lifted its five - year ban on the \textcolor{red}{return} of Imelda Marcos today and said the widow of the late President Ferdinand Marcos was free to come home from exile in the United States.\\}
    \end{minipage}
In this example, the return of Imelda Marcos is introduced in the subsection ``Return from exile (1991–present)'' of the page ``Imelda Marcos''. However, we only use the first 2,000 characters of the page contents to represent the title ``Imelda Marcos'', which has no information about the return from exile. A document-level representation may be a potential solution for future works.

(ii): Sub-events. Some events are sub-events of other larger events. For example:

    \begin{minipage}{0.93\linewidth}
    {\fontsize{10}{12}\selectfont Stepping in at the 11th hour, Hillary Rodham Clinton \textcolor{red}{will campaign} in Florida on Saturday for her brother, Hugh Rodham, in his bid for a United States Senate seat.\\}
    \end{minipage}
This event is a sub-event of ``1994 United States Senate election in Florida'', which has different event arguments, so the names in the local context do not overlap with the names in the page. 

In this work, we discuss many challenges of the task in different domains, but \modelname~cannot address all of them. We leave them to future works.

\section{Conclusion}

In this work, we formulate Event Linking, a challenging but essential task, with a carefully designed Wikipedia dataset and NYT test set, and propose an event linking model \modelname~for future works.

\section{Limitation}
In this section, we discuss limitations of our work.
\begin{itemize}
    \item We only focus on event linking to English Wikipedia in this work. We leave multilingual event linking to future works.
    \item The performance of \modelname~on hard cases is still low, for example events that only exist in the subsection of Wikipedia page.
    \item In this work, we simply predict all the mentions with a prediction probability that is lower than 50 to ``Nil''. We leave better solutions to future works.
\end{itemize}

\section*{Acknowledgments}

This work was supported by Contract FA8750-19-2-1004 with the US Defense Advanced Research Projects Agency (DARPA), the Oﬃce of the Director of National Intelligence (ODNI), Intelligence Advanced Research Projects Activity (IARPA), via IARPA Contract No. 2019-19051600006 under the BETTER Program, and a Focused Award from Google. Approved for Public Release, Distribution Unlimited. The views and conclusions contained herein are those of the authors and should not be interpreted as necessarily representing the oﬃcial policies, either expressed or implied, of ODNI, IARPA, the Department of Defense, or the U.S. Government. The U.S. Government is authorized to reproduce and distribute reprints for governmental purposes notwithstanding any copyright annotation therein.

\bibliography{anthology,custom,cited,new}
\bibliographystyle{acl_natbib}

\appendix

\section{Implementations}
\label{sec:implement}
We use 4 Nvidia RTX A6000 48GB GPUs for model training and evaluation. For both candidate generation model and candidate ranking model, we train 10 epochs with learning rate $1\mathrm{e}^{-5}$, and use BERT-large-uncased as the pretrained language model \citep{Devlin2019BERTPO}. The maximum tokens of both event mention representation and Wikipedia title representation are 256. Top-K is chosen from [5, 10, 30, 50, 70, 100], and tuned on development data. The Glove version we use is "glove-wiki-gigaword-100". \\\\
The Wikipedia dump version we use is 2020/03/01, which is also released to public with our annotated data. 

\section{Data Annotation}
We require all the annotators from Amazon Mechanical Turk to be English speaker, and with an acceptance rate higher than 95\%. All the annotators are English native speakers and are paid more than 10 US dollars per hour. 

\end{document}